\pdfoutput=1
\documentclass[10pt,twocolumn,letterpaper]{article}

\usepackage{cvpr}
\usepackage{times}
\usepackage{epsfig}
\usepackage{graphicx}
\usepackage{amsmath}
\usepackage{amssymb}
\usepackage{multirow}
\usepackage{algorithm}
\usepackage{tabularx}
\usepackage{subfigure}
\usepackage{algorithmic}
\usepackage{float}

\usepackage[breaklinks=true,colorlinks,bookmarks=false]{hyperref}

\cvprfinalcopy %

\def\cvprPaperID{150} %

\ifcvprfinal\fi
\begin{document}

\title{Projection \& Probability-Driven Black-Box Attack}

\author{
    Jie Li$^{1}$,
    Rongrong Ji$^{1}$\thanks{Corresponding author.},
    Hong Liu$^{1}$,
    Jianzhuang Liu$^{2}$,
    Bineng Zhong$^{3}$,
    Cheng Deng$^{4}$,
    Qi Tian$^{2}$
    \\
    $^{1}$Department of Artificial Intelligence, School of Informatics, Xiamen University, \\
    $^{2}$Noah's Ark Lab, Huawei Technologies
    $^{3}$Huaqiao University
    $^{4}$Xidian University
    \\
    {\tt\small lijie.32@outlook.com,}
    {\tt\small rrji@xmu.edu.cn,}
    {\tt\small lynnliu.xmu@gmail.com,}
    {\tt\small liu.jianzhuang@huawei.com,}\\
    {\tt\small bnzhong@hqu.edu.cn,}
    {\tt\small chdeng.xd@gmail.com,}
    {\tt\small tian.qi1@huawei.com,}
}

\maketitle
\thispagestyle{empty}

\begin{abstract}
    Generating adversarial examples in a black-box setting
    retains a significant challenge with vast practical application prospects.
    In particular, existing black-box attacks suffer from the need for excessive queries, as it is non-trivial to find an appropriate direction to optimize in the high-dimensional space.
    In this paper, we propose Projection \& Probability-driven Black-box Attack (PPBA) to tackle this problem by reducing the solution space and providing better optimization.
    For reducing the solution space,
    we first model the adversarial perturbation optimization problem as a process of recovering frequency-sparse perturbations with compressed sensing,
    under the setting that random noise in the low-frequency space is more likely to be adversarial.
    We then propose a simple method to construct a low-frequency constrained sensing matrix, which works as a plug-and-play projection matrix to reduce the dimensionality.
    Such a sensing matrix is shown to be flexible enough to be integrated into existing methods like NES and Bandits$_{TD}$.
    For better optimization,
    we perform a random walk with a probability-driven strategy,
    which utilizes all queries over the whole progress to make full use of the sensing matrix for a less query budget.
    Extensive experiments show that our method requires at most $24\%$ fewer queries with a higher attack success rate compared with state-of-the-art approaches.
    Finally, the attack method is evaluated on the real-world online service, i.e., Google Cloud Vision API, which further demonstrates our practical potentials.
    \footnote{The code for reproducing our work is available at \url{https://github.com/theFool32/PPBA}}
\end{abstract}

\section{Introduction}

While deep neural networks (DNNs) have proven their dominant performance on a wide range of computer vision tasks,
they are shown to be
vulnerable to adversarial examples~\cite{Li_2019_ICCV,szegedy2013intriguing,xie2017adversarial}.
In such a scenario, the imperceptible perturbations added to input samples can mislead the output of DNNs,
which has raised serious security concerns in the literature~\cite{carlini2016hidden,EykholtEF0RXPKS18,sharif2016accessorize}.

Adversarial attacks can be generally categorized into the \textit{white-box} attack and \textit{black-box} attack.
In white-box attacks, the adversary has the full knowledge of the victim model including network architecture and parameters,
and
can efficiently achieve an almost $100\%$ attack success rate within a few iterations guided by the gradient descent~\cite{carlini2017towards,goodfellow2014explaining,madry2017towards,moosavi2016deepfool}.
However, white-box attacks are less practical for commercial systems like Google Cloud Vision API, where the model is inaccessible.

To this end, the black-box attacks, including transfer-based attacks and query-based attacks, are more practical where the adversary is only able to craft model inputs and obtain corresponding outputs.
Transfer-based attacks~\cite{dong2019evading,pengcheng2018query,papernot2017practical} adopt the adversarial perturbation crafted from a surrogate white-box model and transfer it to the black-box victim model.
They require less time consumption, but suffer from low attack performance, since the target model may be very different from the surrogate.
To achieve a high attack success rate, recent works~\cite{chen2019hopskipjumpattack,chen2017zoo} queried iteratively to estimate the gradients and then perform the white-box attacks, or approach the decision boundary first and then wander along it.
Due to the high-dimensional input space, it is hard to find a feasible direction to optimize, which results in numerous queries and a high cost of time and money.
Some extra efforts have been put on reducing the dimension of solution space, like utilizing latent variable space of autoencoder~\cite{tu2019autozoom} or adopting low-resolution images~\cite{ilyas2018prior}.
Essentially, these methods reduce the solution space from the perspective of the spatial domain,
and the reduction of dimension is limited since images with quite low-resolution will be unserviceable,
which makes these methods still inefficient.

In this paper, we propose Projection \& Probability-driven Black-box Attack (PPBA) towards achieving a high attack success rate with few queries.
Optimization in a high dimension is difficult and urges for a smaller solution space~\cite{datar2004locality,liu2018ordinal}.
On the other hand, some recent works~\cite{guo2018low,sharma2019effectiveness} have experimentally verified that adversarial perturbations tend to lie in the low-frequency space, which is a subspace of the original solution space.
These both motivate us to form a smaller search space from the frequency perspective.
Considering that, we first reduce the query number via reducing the solution space with a low-frequency constrained projection matrix.
In particular, we view this problem as recovering adversarial perturbations using compressed sensing with a sensing matrix.
This sensing matrix can be crafted by applying the inverse Discrete Cosine Transform (DCT)~\cite{ahmed1974discrete} on the standard basis and selecting the low-frequency parts.
The sensing matrix is plug-and-play as a projection matrix.
With the elaborate sensing matrix, we reduce the dimension of the solution space from that of the image space
(\emph{e.g.}, $224 \times 224 \times 3=150,528$) to a very small one (\emph{e.g.}, $1,500$).
Based on this sensing matrix, we then propose a more suitable attack strategy driven by probability.
We care merely about the direction of each dimension, upon which we quantize the value of every iteration into a triplet.
Then, a probability-driven strategy is kicked in to take advantage of information throughout the iteration process to perform a random walk optimization.

Extensive experiments show the efficiency of the proposed PPBA method.
By integrating the proposed low-frequency sensing matrix into various existing methods,
we verify that it is flexible enough, which can reduce $9.6\%$ queries with a higher attack success rate for VGG-16~\cite{simonyan2014very} on ImageNet~\cite{deng2009imagenet}.
PPBA further improves the performance over the state-of-the-art methods~\cite{guo2019simple,ilyas2018black,ilyas2018prior} with at least $11\%$ fewer queries for Inception v3~\cite{szegedy2016rethinking} on ImageNet.
Finally, we evaluate PPBA on the real-world image classifier, \emph{i.e.}, Google Cloud Vision API, and show that our method can efficiently corrupt it with an $84\%$ success rate.

Concretely, the contributions of this work are as follows:
\begin{itemize}
    \item We view generating adversarial perturbations as recovering sparse signals and propose a low-frequency sensing matrix to efficiently reduce the dimension of the solution space.
        The sensing matrix is plug-and-play and can be integrated into existing methods functioned as a projection matrix.
    \item Based on this projection matrix,  a probability-driven attack method is proposed, which suits the sensing matrix more and makes the best use of the information throughout the whole iteration process.
    \item The proposed PPBA method achieves higher performance on different neural networks~\cite{he2016deep,simonyan2014very,szegedy2016rethinking} pre-trained on ImageNet~\cite{deng2009imagenet}, compared with state-of-the-art methods~\cite{guo2019simple,ilyas2018black,ilyas2018prior}, and can fool real-world systems efficiently.
\end{itemize}

\section{Related Work}

\subsection{White-Box Attacks}
The adversary under the white-box settings has full knowledge of the victim model.
Szegedy \emph{et al.}~\cite{szegedy2013intriguing} first demonstrated that intentionally perturbated images, \emph{e.g.} by adding quasi-imperceptible adversarial perturbations, can fool neural networks.
These adversarial perturbations can be crafted with box-constrained L-BFGS~\cite{liu1989limited}.
Subsequently, various methods have been proposed to generate such perturbations.
For example,
Goodfellow \emph{et al.}~\cite{goodfellow2014explaining,kurakin2016adversarial} took a linear view of adversarial examples and proposed fast ways of generating them in one step or iteratively.
Moosavi-Dezfooli \emph{et al.}~\cite{moosavi2016deepfool} attempted to find adversarial examples from the decision boundary.
Carlini \emph{et al.}~\cite{carlini2017towards} compared different objective functions
and proposed a powerful C\&W attack method.
Note that these methods perform optimization with the gradient information, which cannot be applied to black-box attacks directly.

\subsection{Black-Box Attacks}
White-box attacks are unrealistic for many real-world systems, where neither model architectures nor parameters are available.
Under this scenario, black-box attacks are necessary.
In black-box attacks, the adversary is unable to access the target victim model, and only the model inputs and its corresponding outputs can be fetched.
In this paper, we assume that the outputs include prediction confidences since it is a common setting for popular online systems, \emph{e.g.}, Google Cloud Vision, Clarifai, and Microsoft Custom Vision.
There are two types of black-box attack methods, \emph{i.e.}, transfer-based attacks and query-based attacks:

\textbf{Transfer-Based Attacks.}
Since models trained on the same dataset may share similar decision boundaries, adversarial examples can transfer across models to some degree.
Considering that, the adversary performs a standard white-box attack on accessible local models to construct adversarial examples, which are expected to be transferred to the inaccessible target model.
One type of such attack assumes that the local model and target model are trained with data from similar distributions, and no query on the target model is needed~\cite{liu2019universal,moosavi2017universal,szegedy2013intriguing}.
Another type of such attack is to distill the target model with a surrogate model~\cite{pengcheng2018query,papernot2017practical}, which requests a large number of queries to train the local model and is thus inefficient.
Although there exist many works focusing on improving the transferability of adversarial examples~\cite{dong2019evading,xie2019improving,zhou2018transferable}, the attack success rates of transfer-based attacks are still less competitive to query-based attacks.

\textbf{Query-Based Attacks.}
Query-based attacks define an objective function and update the perturbation iteratively to optimize this function.
Each iteration requires one or more queries to determine the next step.
Authors in~\cite{dong2019evading,su2019one} constructed the adversarial perturbations with an evolutionary algorithm.
The efficiency of evolutionary algorithms is highly dependent on the dimension of the inputs and the size of the solution space, which makes these algorithms time-consuming.
The authors in~\cite{brendel2017decision,chen2019hopskipjumpattack} proposed the decision-based attack that initiates perturbations from a target image or with a large norm to guarantee adversarial and then reduces the norm iteratively along the decision boundary.
Despite a high success rate, this kind of method requires a large number of queries wandering along the decision boundary as the boundary can be potentially complex.
Another mainstream of query-based attacks is to estimate the gradients and then perform the white-box attack.
Chen \emph{et al.}~\cite{chen2017zoo} proposed the ZOO (zeroth-order optimization) attack that adopts the finite-difference method with dimension-wise estimation to approximate the gradient values.
It takes $2d$ queries in each iteration, where $d$ is the dimension of the input image ($d$ can be more than $150,000$).
Bhagoji \emph{et al.}~\cite{bhagoji2017exploring} attempted to reduce the query budget in each iteration via random grouping or PCA components mapping.
Tu \emph{et al.}~\cite{tu2019autozoom} utilized a pretrained autoencoder and optimized the perturbations in the latent space.
Instead of using the finite-difference method, Ilyas \emph{et al.}~\cite{ilyas2018black} proposed the NES attack that adopts the natural evolution strategy to estimate gradients with random vectors.
The Bandits$_{TD}$ is further proposed in~\cite{ilyas2018prior}, which incorporates time and data-dependent information with the bandit theory to reduce the query cost.
Guo \emph{et al.}~\cite{guo2019simple} proposed the SimBA-DCT that adds or subtracts random vectors iteratively from a set of orthonormal vectors to craft adversarial examples.
Considerable query cost is reduced by the aforementioned methods, which is however still far from satisfactory.

\section{The Proposed Method}

The large solution space for black-box and the inefficient optimization retain as two key bottlenecks for existing black-box attack methods.
To solve these two issues, we first reduce the solution space from its original dimension with a low-frequency constrained sensing matrix, as detailed in Sec.\,\ref{sec_project}.
Then, to further reduce the query cost, we propose a novel weighted random walk optimization based on the sensing matrix,
as described in Sec.\,\ref{sbsec_probability_driven_optimization}.

\subsection{Preliminaries}
Given a deep neural network classifier $f: {[0,1]}^d \rightarrow \mathbb{R}^K$ that maps the input image $x$ of $d$ dimensions into the confidence scores of $K$ classes,
we define $F(x)=\arg\max_{k} f(x)_k$ as the function that outputs the predicted class.
The goal of adversarial attack against classification is to find a perturbation $\delta \in \mathbb{R}^d$ that satisfies:
\begin{align}\label{orginal_objective}
   F\big(\Pi_{Img}(x+\delta)\big) \neq F(x), \text{ s.t. } \|\delta\|_p < \epsilon,
\end{align}
where $\Pi_{Img}(\cdot)=clip(\cdot, 0, 1)$\footnote{For simplicity, we will omit it in what follows.} is a projection function that projects the input into the image space, \emph{i.e.}, $[0,1]^d$, and $\epsilon$ is a hyper-parameter to make the perturbation invisible via restricting the $l_p$-norm.
To achieve the goal, we adopt the widely used objective function termed C\&W loss~\cite{carlini2017towards}:
\begin{align}\label{loss_function}
\min\limits_{\|\delta\|_{p} < \epsilon} L(\delta) = {[{f(x+\delta)}_t - \max\limits_{j\neq t}\big({f(x+\delta)}_{j})]}^{+},
\end{align}
where ${[\cdot]}^{+}$ denotes the $\max(\cdot, 0)$ function,  $t$ is the label of the clean input.
For iterative optimization methods, to guarantee the constraint $\|\delta\|_p < \epsilon$, another projection function is needed after each update of $\delta$. For instance, for $l_2$-norm, a projection function $\Pi_{2}(\delta, \epsilon)=\delta * min(1, \epsilon / \|\delta\|_{2})$ should be applied in each step.

\subsection{Low-Frequency Projection Matrix}\label{sec_project}
\subsubsection{Perspective from Compressed Sensing}\label{sbsec_compressed_sensing}

Recent works have discovered that adversarial perturbations are biased towards the low frequency information~\cite{guo2018low,sharma2019effectiveness}.
Suppose there exists an optimal low-frequency perturbation $\delta^*$ that is sparse in the frequency domain for Eq.\,(\ref{orginal_objective}).
Thus,
$\Psi \delta^*$ should be a sparse vector, where $\Psi \in \mathbb{R}^{d\times d}$ is the transform matrix of DCT that maps a vector from the time/spatial domain to the frequency domain and satisfies $\Psi\Psi^T = \Psi^T\Psi = I_{d \times d}$, where $I_{d\times d}$ is the indentify matrix..
According to the compressed sensing theory~\cite{candes2006stable, donoho2006compressed}, we can recover the sparse vector with a measurement matrix $\Phi \in \mathbb{R}^{m\times d} (m\ll d)$ and the corresponding measurement vector $z\in \mathbb{R}^m$ by:
\begin{align}\label{objective_2}
    &\min  \|\Psi\delta^*\|_2, \nonumber \\
    \text{s.t.}\ &z =A \delta^*= \Phi \Psi \delta^*, \nonumber
    \\ &F(x+\delta^*) \neq F(x),
\end{align}
where $A = \Phi\Psi \in \mathbb{R}^{m\times d}$ is the sensing matrix.

The measurement matrix $\Phi$ can be further simplified as $\Phi=[\Phi_m,\textbf{0}], (\Phi_m \in \mathbb{R}^{m\times m}, \textbf{0} \in \mathbb{R}^{m\times (d-m)})$, to suppress high frequency, considering that $\delta^*$ is biased to low frequency.
Note that orthogonal matrices do not change the norm of a vector after transforming it, which also guarantees the restricted isometry property~\cite{candes2008restricted,candes2006near} required by the compressed sensing theory.
Therefore, we directly set $\Phi_m$ as an orthogonal matrix, and recover the perturbation $\delta^*$ with simple matrix multiplication as:
\begin{align}\label{transform}
    z &= \Phi\Psi\delta^*, \nonumber \\
    \Phi^Tz &\approx \Psi\delta^*, \nonumber \\
    \Psi^T\Phi^Tz &= A^Tz \approx \delta^*.
\end{align}
Finally, Eq.\,(\ref{objective_2}) can be rewritten as:
\begin{align}\label{objective_final}
    &\min \|z\|_2, \nonumber \\
    \text{s.t.}\ &F(x+A^Tz) \neq F(x).
\end{align}
As a result, we only need to perform the optimization in the $m$-dimensional space instead of the $d$-dimensional one ($m \ll d$),
which results in a smaller solution space and the higher optimization efficiency.

\subsubsection{Perspective from Low Frequency}\label{sbsec_low_frequency}
From another perspective, we discover that the measurement vector $z$ optimized in Eq.\,(\ref{objective_final}) has its physical meaning.
The optimal perturbation vector $\delta^*$ can be linearly represented by the discrete cosine basis as below:
\begin{align}\label{decompose}
    \delta^* = \sum_j \alpha_j \omega_j,
\end{align}
where $w_{j}$ is a discrete cosine basis vector, and $a_{j}$ is the corresponding coefficient.
Note that $w_{j}$ contains the specific frequency information, we can also view Eq.\,(\ref{decompose}) as decomposing $\delta^{*}$ into the sum of different frequency vectors and the $\alpha_{j}$ is the corresponding amplitude.
Since the vector $\omega_j$ can be easily crafted via applying inverse DCT on one of the standard basis vectors,
we then rewrite Eq.\,(\ref{decompose}) into the form of matrix multiplication as below:
\begin{align}\label{decompose_matrix}
    \delta^* = \Omega\alpha = \Psi^T Q \alpha,
\end{align}
where $\Omega$ is a matrix formed by the frequency vectors $\omega_j$ as its columns,
$\Psi^T$ is the transform matrix of inverse DCT as mentioned before, $Q \in \mathbb{R}^{d \times m}$ is a submatrix subsampled from the standard basis $I_{d \times d}$ for low frequency, and $\alpha$ is the amplitude vector.

Comparing Eq.\,(\ref{decompose_matrix}) with Eq.\,(\ref{transform}), it is inspiring to find that:
\begin{equation}
        \left\{
    \begin{aligned}
    & \delta^* = \Psi^T Q \alpha, \\ \nonumber
    & \delta^* \approx \Psi^T \Phi^T z,
    \end{aligned}
    \right.
    \Rightarrow \Psi^T Q \alpha \approx \Psi^T \Phi^T z.
\end{equation}
Since $Q$ is orthogonal, it suggests a simple and efficient way to construct the sensing matrix $A$ by applying inverse DCT to the standard basis\footnote{For 2D images, we utilize 2D IDCT, \emph{i.e.}, utilize 1D IDCT twice.}, and the measurement vector $z$ is just the amplitude $\alpha$ in Eq.\,(\ref{decompose_matrix}).

\subsection{Probability-Driven Optimization}\label{sbsec_probability_driven_optimization}
As discussed in Sec.\,\ref{sbsec_low_frequency}, the measurement vector $z$ can be viewed as the amplitude.
Therefore, the change of $z$ in each iteration can be simplified by a triplet $\{-\rho,0,\rho\}$, denoting decreasing the corresponding amplitude value by $\rho$, keeping it, and increasing it by $\rho$, respectively.
Then the choice space of the iteration step is further restricted.
Based on this setting, a random walk optimization is further adopted, which chooses steps randomly and moves when the step makes the loss descend.

To achieve better performance, instead of adopting the random steps, we make the best of information in the past by assuming that the directions of steps in the past are capable of guiding the choice of the current step to a certain degree.
We rewrite the objective function in Eq.\,(\ref{loss_function}) as:
\begin{equation}
    \label{eq_our_loss}
    \small
    L(z, A) = {[{f(x+A^Tz)}_t - \max\limits_{j\neq t}\big({f(x+A^Tz)}_j\big)]}^{+},
\end{equation}
where an iterative optimization method like random walk can be applied.
In particular, after defining $\Delta z$ as the change of $z$ in an iteration of random walk, a confusion matrix is calculated for each dimension $\Delta z_j$ of $\Delta z$ as:
\begin{table}[H]
    \centering
    \footnotesize
    \resizebox{0.3\textwidth}{!}{
    \begin{tabular}{l|l|l|l}
    \hline
     & $-\rho$ & 0 & $\rho$ \\
     \hline
     \# effective steps & $e_{-\rho}$ & $e_{0}$ & $e_{\rho}$ \\
     \# ineffective steps & $i_{-\rho}$ & $i_{0}$ & $i_{\rho}$ \\
\hline
\end{tabular}
    }
\end{table}
\noindent where $e_{-\rho}$ means the number of times the loss function (\emph{e.g.}, Eq.\,(\ref{loss_function})) descends when $\Delta z_j = -\rho$, and $i_{-\rho}$ means the number of times the loss function keeps still or ascends when $\Delta z_j = -\rho$.
We calculate the effective rate for every possible value with:
\begin{equation}
    \small
    P(\text{effective}|\Delta z_j=v) = \frac{e_{v}}{e_{v}+i_{v}},
    \text{ for } v \in \{-\rho,0,\rho\},
\end{equation}
and sample $\Delta z_j$ with a probability as:
\begin{align}\label{eq_probability}
    P(\Delta z_j=v) = \frac{P(\text{effective}|\Delta z_j=v)}{\sum_{u} P(\text{effective}|\Delta z_j=u)},  \nonumber \\
    \text{for } v \text{ and } u \in \{-\rho,0,\rho\}.
\end{align}
Therefore, when $\Delta z_j=v$, an effective query step increases the value of $e_v$ along with $P(\text{effective}|\Delta z_j=v)$, which results in an increase in $P(\Delta z_j=v)$, and an ineffective one vice versa.
\begin{algorithm}[!t]
    \caption{Projection \& Probability-Driven Black-Box Attack}\label{alg_proposed}
    \renewcommand{\algorithmicrequire}{\textbf{Input:}}
    \renewcommand{\algorithmicensure}{\textbf{Output:}}
    \begin{algorithmic}[1]
        \REQUIRE{}
        Input image $x$, maximum number of queries $max\_iter$.
    \ENSURE{}
    Perturbation vector $\delta$.
    \STATE{} Initialize $z \leftarrow \textbf{0} \in \mathbb{R}^{m}$, confusion matrices with all elements of $1$, and $j\leftarrow 0$
    \STATE{} Construct sensing matrix $A$ via applying IDCT to the submatrix of $I_{d \times d}$
    \FOR{$j < max\_iter$}
    \STATE{Generate $\Delta z$ according to Eq.\,\ref{eq_probability}}
    \IF{$L\big(\Pi_2(z+\Delta z), A\big) < L(z, A)$ for $L$ in Eq.\,(\ref{eq_our_loss})}
    \STATE{$z \leftarrow \Pi_2(z + \Delta z)$}
    \ENDIF{}
    \IF{$L(z, A) <= 0$}
    \STATE{break.}
    \ENDIF{}
    \STATE{Update the confusion matrices accordingly}
    \STATE{$j \leftarrow j+1$}
    \ENDFOR{}
    \STATE{return $\delta = A^Tz$}
    \end{algorithmic}
\end{algorithm}

We prove that with $T$ iterations, the norm of perturbation $\delta$ is bounded by:
\begin{align}
    \|\delta\|_2 & = \|A^Tz\|_2
    = tr\big((A^T z)^TA^T z\big) \nonumber \\
        &= tr(z^TAA^T z)
        = tr(z^Tz) = \|z\|_2  \nonumber \\
        & =\|\sum_j \Delta z^j\|_2 \leq \|T \times \vec{\rho}\|_2   \nonumber \\
        & = \sqrt{m} \times T \times \rho,
\end{align}
where $\Delta z^j$ is the $\Delta z$ of the $j$-th iteration and $\vec{\rho}$ is a vector with all elements of value $\rho$.
Despite the above prove,
we still utilize the projection function $\Pi_2(\cdot)$ to keep the norm constraint.
For each iteration,
we evaluate whether $\Pi_2(z+\Delta z)$ can successfully decrease the objective function and update the confusion matrices. We accept the $\Delta z$ and update $z=\Pi_2(z+\Delta z)$ only if it succeeds.
The above process repeats until we find an adversarial perturbation or meet the maximum number of iterations.
We refer to this method as Projection \& Probability-driven Black-box Attack (PPBA), and
the detailed algorithm is provided in Alg.\,\ref{alg_proposed}.

\section{Experiments}

\subsection{Experimental Setups}
\textbf{Datasets and Victim Models.}
We evaluate the effectiveness of our proposed PPBA along with baselines on ImageNet~\cite{deng2009imagenet}.
For each evaluation, we sample $1,000$ images (one image per class) randomly from the validation set for evaluation.
For the victim models, we choose the widely-used networks pre-trained on ImageNet, \emph{i.e.}, ResNet50~\cite{he2016deep}, VGG-16~\cite{simonyan2014very}, and Inception V3~\cite{szegedy2016rethinking}.
Considering the charge cost of Google API (\$$1.50$ for $1,000$ queries), we randomly select $50$ images to evaluate the results on the Google Cloud Vision API\footnote{\url{https://cloud.google.com/vision/docs/drag-and-drop}}.

\textbf{Evaluation Metric.}
Restricting the norm of the resulting perturbation,
there are two aspects to evaluate black-box adversarial attacks:
How often a feasible solution can be found, and how efficient the optimization method is.
The attack success rate can quantitatively represent the first one.
We define a successful attack for ImageNet as the one that changes the top-1 predicted label within the maximum queries.
For the second one, the average number of queries (abbreviated to average queries) can give a rough sense.
We report the average queries on both success samples and all samples.
The average on success samples denotes how many queries are needed to successfully perturb an input, which is more useful.
However, it is strongly connected to the success rate and thus gives a false sense for low success rate attack methods.
We thereby report the average on all samples as a supplement.
Considering that samples with a large number of queries have large impacts on the average value,
we further depict the curve of the success rate versus the number of queries and calculate the area under the curve (AUC) for a better comparison.

\textbf{Compared Methods and Settings.}
We mainly compare our proposed PPBA with NES~\cite{ilyas2018black},
Bandits$_T$~\cite{ilyas2018prior} (Bandits with the time-dependent prior),
Bandits$_{TD}$~\cite{ilyas2018prior} (Bandits with the time and data-dependent prior), and SimBA-DCT~\cite{guo2019simple}.
We evaluate the performance of the baselines with the source code released by the authors\footnote{\url{https://github.com/MadryLab/blackbox-bandits}, \url{https://github.com/cg563/simple-blackbox-attack}}, and use the default parameters setting in their papers.
We also perform the random walk with each step uniformly sampled from the triplet space to test the efficiency of our proposed probability-driven strategy.
We name this kind of attack as \textit{Projection \& Random walk Black-box Attack} (PRBA).
Following the settings in~\cite{ilyas2018prior},
we set the maximum $l_2$-norm for perturbations to $5$, and the maximum $l_{\infty}$-norm to $0.05$.
Since $10,000$ is a huge number in reality, we set the maximum number of queries to $2,000$ instead, and set $\rho$ to $0.01$.

\subsection{On the Perturbation $\delta^*$ and Dimension $m$}
Before evaluating the effectiveness of our proposed method,
we first verify that the perturbation $\delta^*$ exists under the sensing matrix setting,
and determine the value of dimension $m$ experimentally.
In this experiment, we utilize another $100$ images randomly sampled from ImageNet for validation.

\subsubsection{On the Existence of Perturbation $\delta^*$}
To verify that the existence of adversarial perturbation $\delta^*$ in the low-frequency space constrained by our sensing matrix, we first perform the white-box attack using the BIM method~\cite{kurakin2016adversarial} with/without the sensing matrix $A$.
The results are shown in Tab.\,\ref{tb_exist}.
All attacks achieve a $100\%$ success rate, which demonstrates that there indeed exist optimal perturbation with the sensing matrix constrained.
Interestingly, we discover that perturbations found with the low-frequency constraint tend to have a much smaller average $l_2$-norm, which is consistent with the results in~\cite{guo2018low}.

\begin{table}[]
    \centering
    \resizebox{0.49\textwidth}{!}{
\begin{tabular}{c|ccc|ccc}
\hline
\multirow{2}{*}{}  & \multicolumn{3}{c|}{Success Rate}  & \multicolumn{3}{c}{Average $L_2$-Norm}          \\ \cline{2-7}
                   & \begin{tabular}[c]{@{}c@{}}R\end{tabular}
                   & \begin{tabular}[c]{@{}c@{}}V\end{tabular}
                   & \begin{tabular}[c]{@{}c@{}}I\end{tabular}
                   & \begin{tabular}[c]{@{}c@{}}R\end{tabular}
                   & \begin{tabular}[c]{@{}c@{}}V\end{tabular}
                   & \begin{tabular}[c]{@{}c@{}}I\end{tabular}
                   \\
                   \hline
BIM     & 100\% & 100\% & 100\% & 4.03 & 3.90 & 5.00 \\
BIM+Sensing Matrix $A$ & 100\% & 100\% & 100\% &  2.06 & 1.70 & 1.86 \\
\hline
\end{tabular}
}
\caption{Results for BIM attack and BIM with our sensing matrix $A$ under the white-box setting. R, V, and I denote ResNet50, VGG-16, and Inception V3, respectively. The 100\% success rate verifies the existence of the adversarial perturbation.}\label{tb_exist}
\vspace{-1em}
\end{table}

\subsubsection{On the Choice of Dimension $m$}
\begin{figure}[!t]
    \begin{center}
        \includegraphics[width=0.49\linewidth]{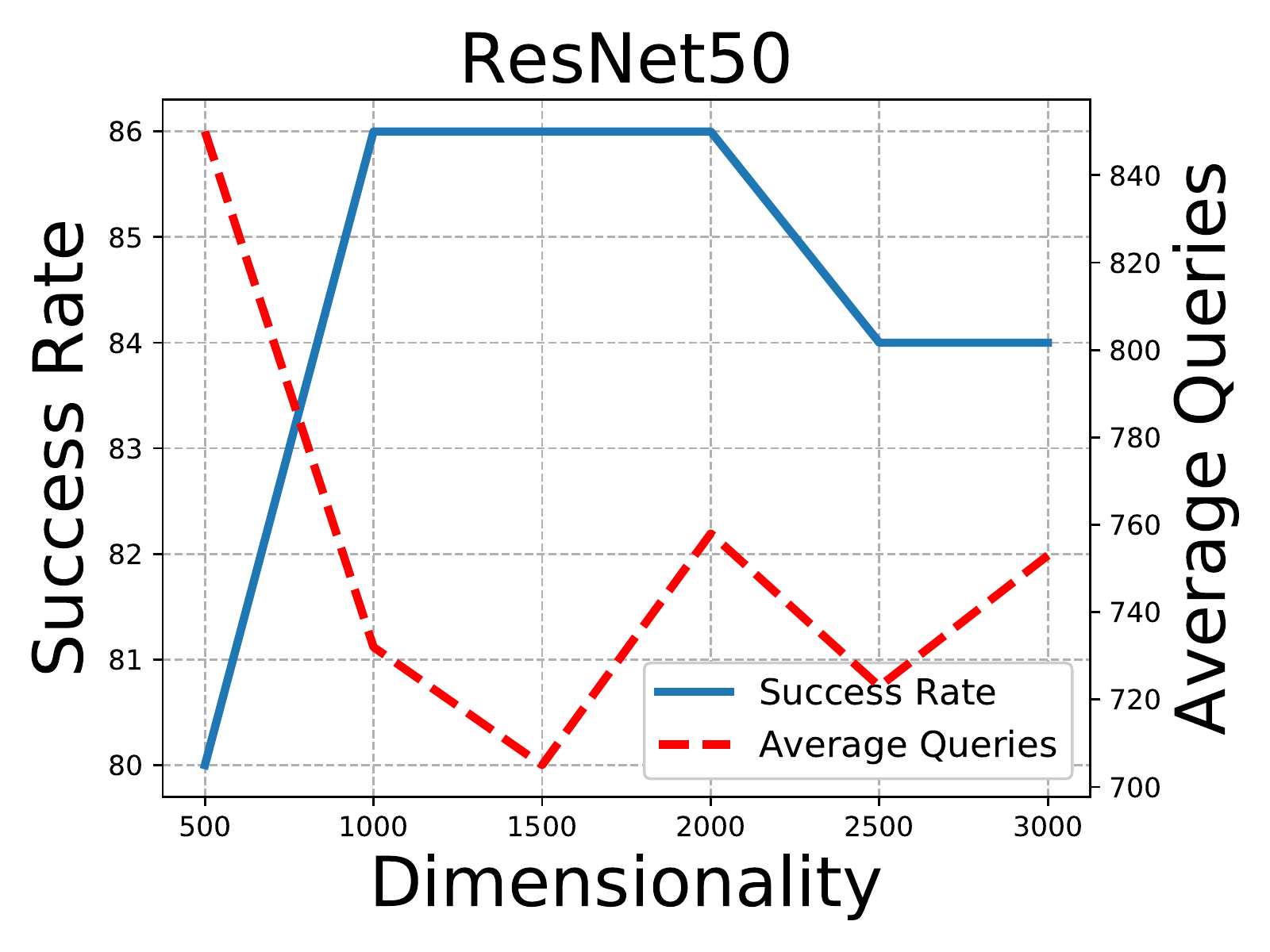}
        \includegraphics[width=0.49\linewidth]{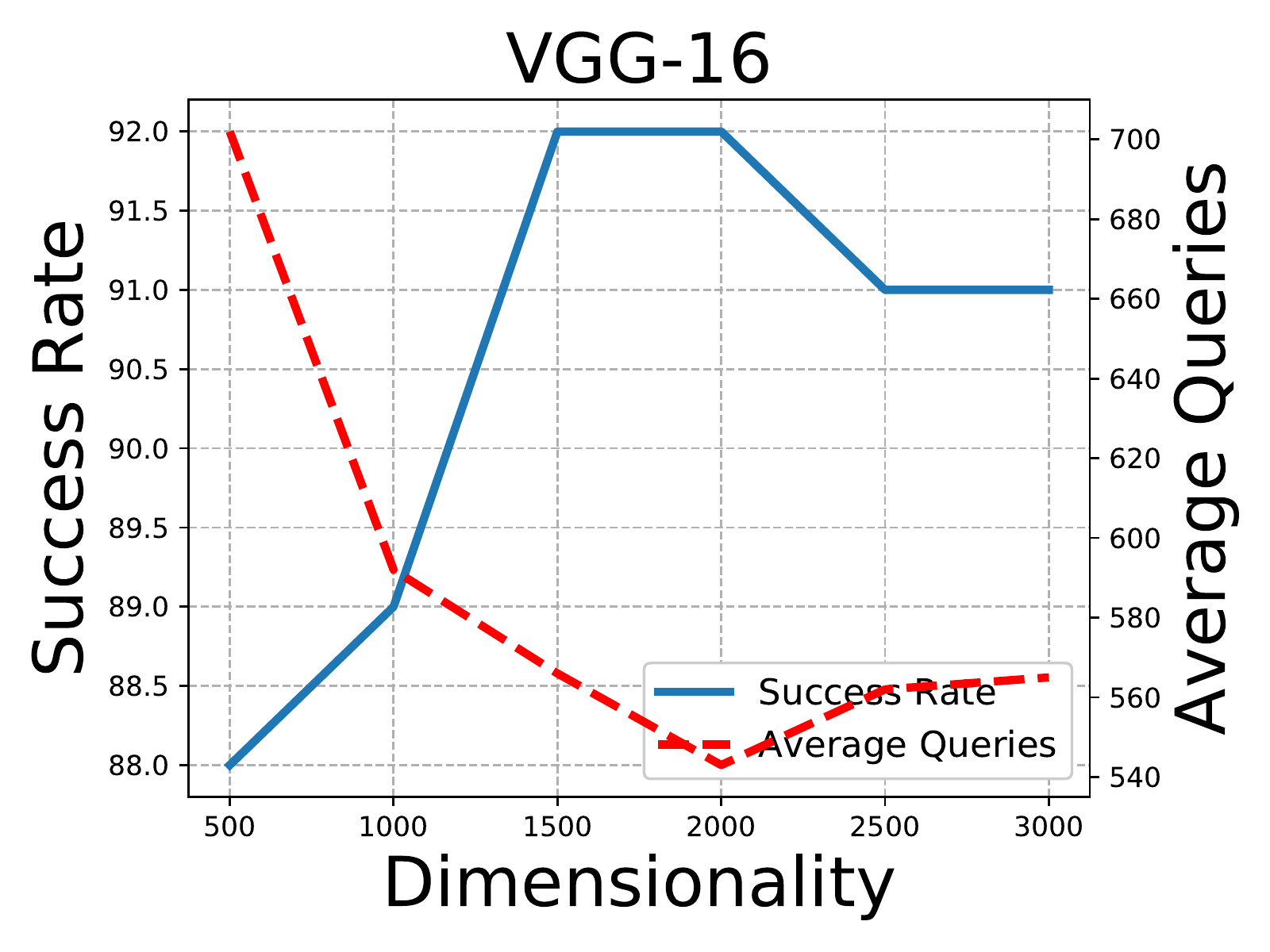}
    \end{center}
\vspace{-1em}
    \caption{The effect of the dimensionality.}
~\label{fg_dim}
\vspace{-1em}
\end{figure}

For determining the dimension of the measurement vectors,
the Johnson–Lindenstrauss lemma~\cite{johnson1984extensions} suggests that for a group of $n$ points in $\mathbb{R}^d$,
there exists a linear map $g:\mathbb{R}^d \rightarrow \mathbb{R}^m,\ m > 8\ln(n)/\epsilon^2$ that keeps the distance between these points:
\begin{equation}
    \small
    (1-\epsilon) \|u-v\|^2 \leq \|g(u)-g(v)\|^2 \leq (1+\epsilon)\|u-v\|^2,
\end{equation}
where $u$ and $v$ are two points, and $\epsilon \in (0,1)$ is a parameter controlling the quality of the projection.
However, since it is hard to count the number of points in the low-frequency space,
we leave the theoretical discovery of the value of $m$ in our future work.
Instead, we choose it experimentally in this paper.
As depicted in Fig.\,\ref{fg_dim}, we evaluate the success rate and average queries by setting different values of $m$ for ResNet50 and VGG-16.
Intuitively, a moderate value is needed since a small dimension degenerates the representation ability and a large dimension enlarges the search space.
Consistent with our intuition, the algorithm is hard to find a satisfying solution with a small dimension, which results in a poor success rate and more queries.
With a large dimension and a large solution space, the algorithm needs more steps to find the optimal solution.
As a result, we set $m$ to $1,500$, $2,000$ and $4,000$ for ResNet50, VGG-16 and Inception V3, respectively.

\subsection{On the Effect of the Sensing Matrix}
As aforementioned, the sensing matrix we design can be integrated into existing methods.
We evaluate the performance after plugging it into NES and Bandits$_T$.
Note that the data-dependent prior~\cite{ilyas2018prior} used in Bandits$_{TD}$ is also a kind of dimensionality reduction (as reduced from the original image space to a low-resolution space with the dimension of $50\times 50\times 3=7,500$),
we compare the performance of our sensing matrix with it.
Another projection matrix, \emph{i.e.}, random Gaussian matrix, is also evaluated.
The quantitative results conducted on $1,000$ randomly selected images can be found in Tab.\,\ref{tb_matrix}.
The results of the random Gaussian matrix show no positive influence on the performance with lower success rates and similar numbers of queries.
On the contrary, the sensing matrix designed with the low-frequency constraint can at most improve the success rate by $27.2\%$, and reduce nearly $26\%$ queries.
Taking Bandits$_{T}$ as baseline,
the data-dependent prior from Bandits$_{TD}$ is an effective method that can reduce $25\%$ queries approximately with a $20.6\%$ success rate improved.
However, our sensing matrix is more effective that can reduce $34\%$ queries with a $24.2\%$ success rate improved.
To explain, the solution space our sensing matrix maps to is much smaller than the one from the data-dependent prior, and the optimal perturbation exists in this space.
Finally, it is worth noting that the low-frequency sensing matrix is plug-and-play, which can improve the performance of other methods efficiently.

\begin{table}[]
    \centering
    \resizebox{0.49\textwidth}{!}{
\begin{tabular}{c|c|c|c|c|c|c}
\hline
\multirow{2}{*}{Methods} & \multicolumn{2}{c|}{ResNet50} & \multicolumn{2}{c|}{VGG-16} & \multicolumn{2}{c}{Inception V3}\\ \cline{2-7}
                        & ASR           & Queries    &  ASR           & Queries         &  ASR           & Queries \\ \hline

NES                & 52.0\%          & 1078/1521                  & 60.7\%          & 1013/1402                  & 26.1\% & 1146/1777 \\
NES+Gaussian       & 50.7\%          & 1035/1511                  & 59.0\%          & 999/1410                   & 25.1\% & 1112/1776 \\
NES+Ours           & \textbf{79.2\%} & \textbf{896}/\textbf{1125} & \textbf{78.3\%} & \textbf{873}/\textbf{1117} & \textbf{48.7\%} & \textbf{958}/\textbf{1493} \\
\hline
Bandits$_{T}$      & 54.1\%          & 719/1306                   & 62.9\%          & 679/1169                   & 34.0\% & 866/1615 \\
Bandits$_{TD}$     & 74.7\%          & 621/970                    & 78.6\%          & 565/871                    & 55.9\% & 701/1274 \\
Bandits$_{T}$+Ours & \textbf{78.3\%} & \textbf{552}/\textbf{867}  & \textbf{79.5\%} & \textbf{474}/\textbf{787}  & \textbf{56.8\%} & \textbf{668}/\textbf{1243} \\

\hline
\end{tabular}
}
\caption{Results of the sensing matrix. Gaussian means the random Gaussian matrix. ASR represents the attack success rate (higher is better). Queries denote the average queries, under which the left number is the amount on success samples and the right number is the amount on all samples (lower is better).}\label{tb_matrix}
\vspace{-1em}
\end{table}

\begin{table*}[]
    \centering
\begin{tabular}{c|ccc|ccc|ccc}
\hline
\multirow{2}{*}{Methods} & \multicolumn{3}{c|}{ResNet50} & \multicolumn{3}{c|}{VGG-16} & \multicolumn{3}{c}{Inception V3}                                                                                                                \\
\cline{2-10}
                         & ASR                           & Queries                     & AUC             & ASR             & Queries                   & AUC             & ASR             & Queries                    & AUC            \\ \hline
NES                      & 52.0\%                        & 1078/1521                   & 481.5           & 60.7\%          & 1013/1402                 & 601.3           & 26.1\%          & 1146/1777                  & 224.3          \\
Bandits$_{T}$            & 54.1\%                        & 719/1306                    & 694.2           & 62.9\%          & 679/1169                  & 831.5           & 34.0\%          & 866/1615                   & 385.7          \\
Bandits$_{TD}$           & 74.7\%                        & 621/970                     & 1030.4          & 78.6\%          & 565/871                   & 1129.3          & 55.9\%          & 701/1274                   & 726.6          \\
SimBA-DCT                & \textbf{87.0\%}               & 604/779                     & 1214.3          & 88.5\%          & 563/722                   & 1271.5          & 61.2\%          & 672/1181                   & 812.6          \\
\hline
PRBA                     & 82.9\%                        & 540/790                     & 1210.6          & 88.5\%          & 489/663                   & 1337.2          & 62.1\%          & 580/1118                   & 882.3          \\
\textbf{PPBA}            & 84.8\%                        & \textbf{430}/\textbf{668}   & \textbf{1331.3} & \textbf{90.3\%} & \textbf{392}/\textbf{548} & \textbf{1451.5} & \textbf{65.3\%} & \textbf{546}/\textbf{1051} & \textbf{948.9} \\

\hline
\end{tabular}
\caption{Results of $l_2$ attack for different methods.}\label{tb_overall}
\vspace{-1em}
\end{table*}

\begin{figure*}
    \begin{center}
    \begin{minipage}[t]{0.33\linewidth}
    \centerline{
    \subfigure[ResNet50]{
    \includegraphics[width=\linewidth]{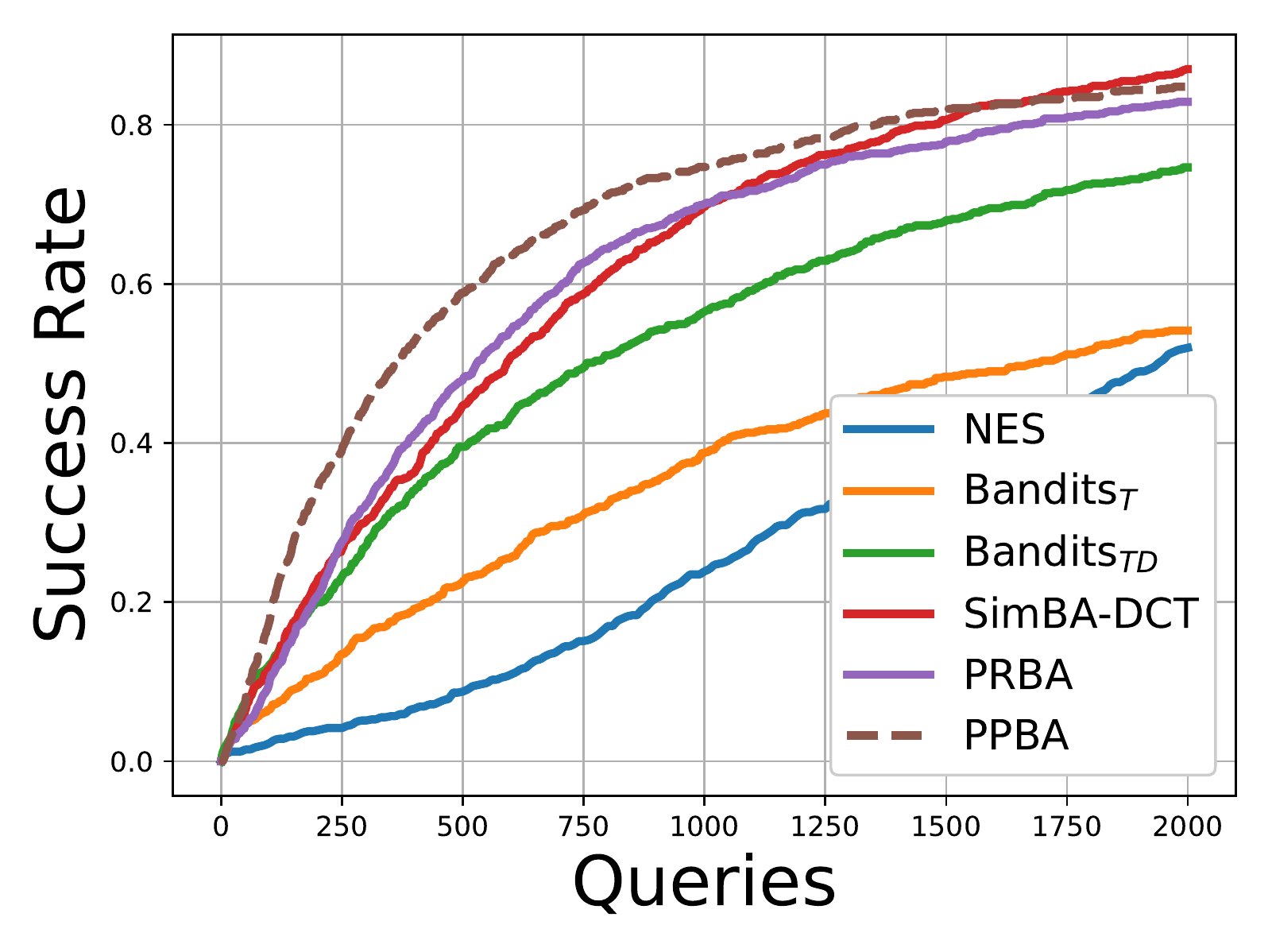}
}
    \subfigure[VGG-16]{
    \includegraphics[width=\linewidth]{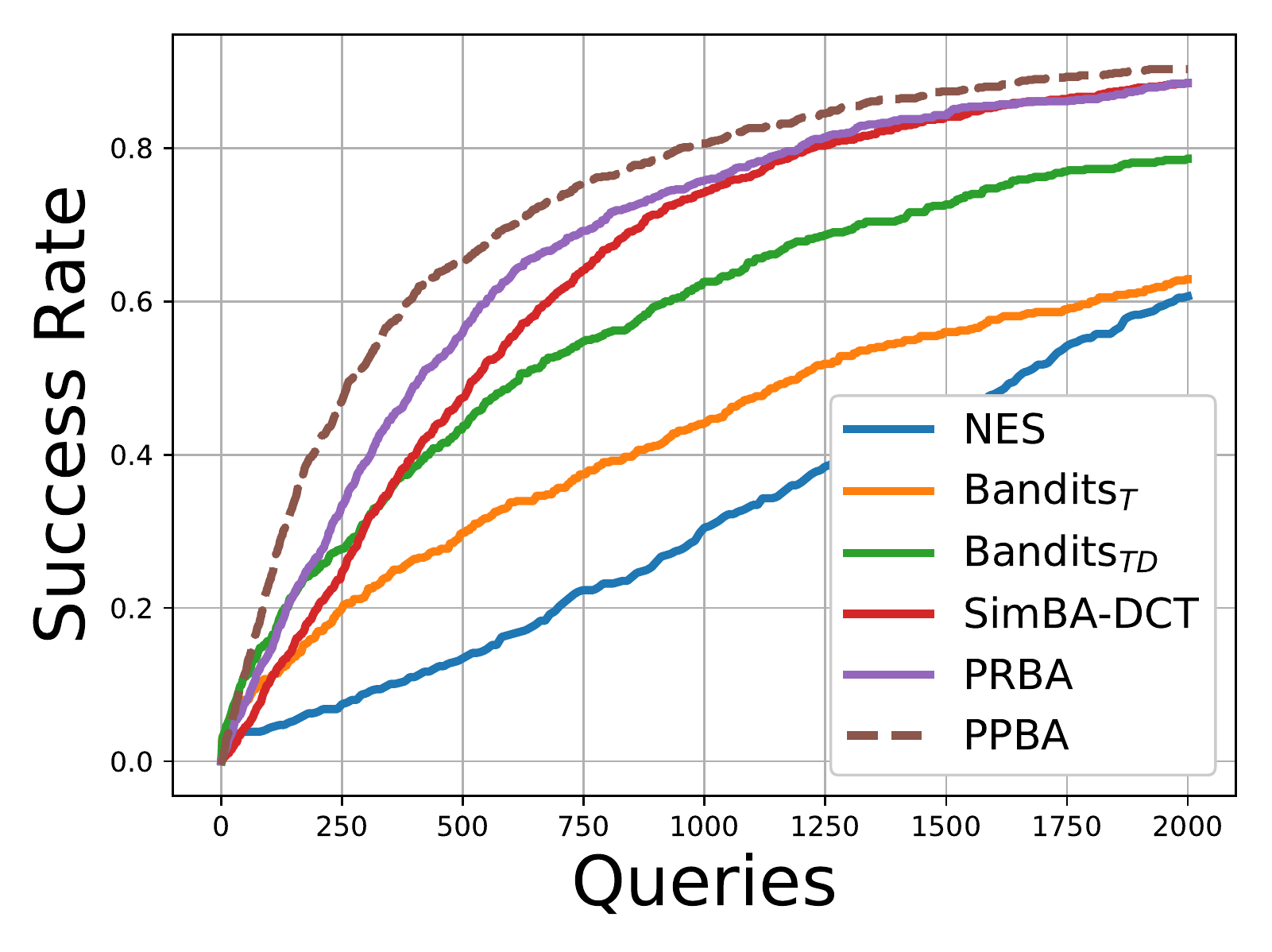}
}
    \subfigure[Inception V3]{
    \includegraphics[width=\linewidth]{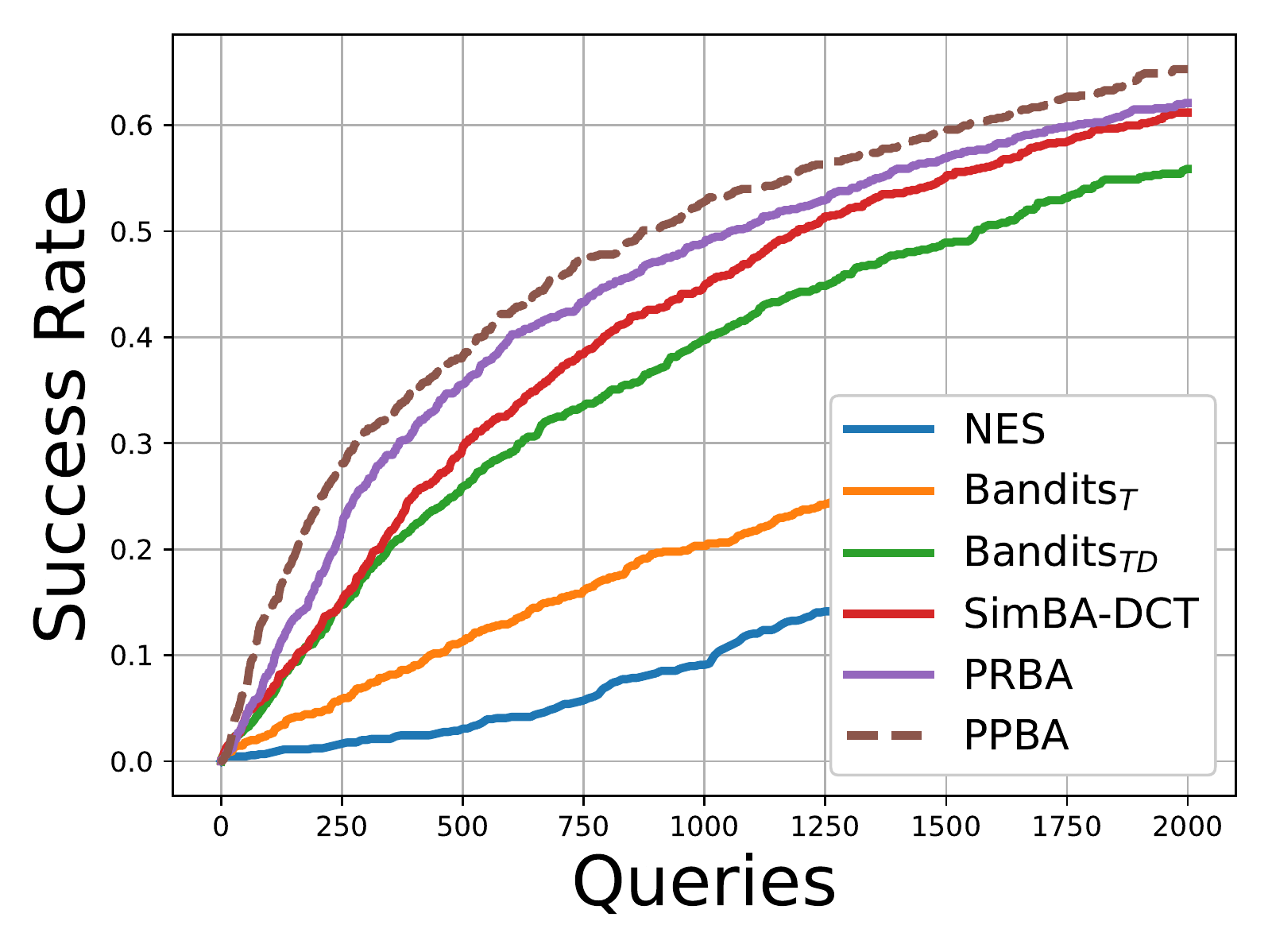}
    }
    }
    \end{minipage}
    \end{center}
     \vspace{-1em}
    \caption{Curves of the attack success rate versus the number of queries for $l_{2}$ attack.}
~\label{fg_curve}
\vspace{-2em}
\end{figure*}

\begin{table}[]
    \centering
    \resizebox{0.49\textwidth}{!}{
\begin{tabular}{c|ccc|ccc|ccc}
\hline
\multirow{2}{*}{Methods} & \multicolumn{3}{c|}{ResNet50} & \multicolumn{3}{c|}{VGG-16} & \multicolumn{3}{c}{Inception V3}                                                                                                                \\
\cline{2-10}
                         & ASR                           & Queries                     & AUC             & ASR             & Queries                   & AUC             & ASR             & Queries                    & AUC            \\ \hline

NES            & 68.7\%          & 867/1222         & 812.7           & 77.7\%          & 745/1026         & 1013.2          & 51.2\%          & 848/1411                   & 606.7          \\
Bandits$_{TD}$ & 84.9\%          & \textbf{409}/648 & 1352.7          & 87.7\%          & \textbf{238}/454 & 1526.6          & 59.9\%          & 592/1162                   & 836.9          \\
SimBA-DCT      & 88.4\%          & 646/797          & 1197.8          & 91.9\%          & 556/667          & 1327.7          & 64.2\%          & 747/1190                   & 804.5          \\
\textbf{PPBA}  & \textbf{96.6\%} & 427/\textbf{481} & \textbf{1519.6} & \textbf{98.2\%} & 337/\textbf{367} & \textbf{1633.1} & \textbf{67.9\%} & \textbf{566}/\textbf{1026} & \textbf{974.2} \\

\hline
\end{tabular}
}
\caption{Results of $l_\infty$ attack for different methods.}\label{tb_inf}
\vspace{-1em}
\end{table}

\begin{figure}[!t]
    \begin{center}
    \begin{minipage}[t]{0.33\linewidth}
        \centerline{
        \subfigure[ResNet50]{
        \includegraphics[width=\linewidth]{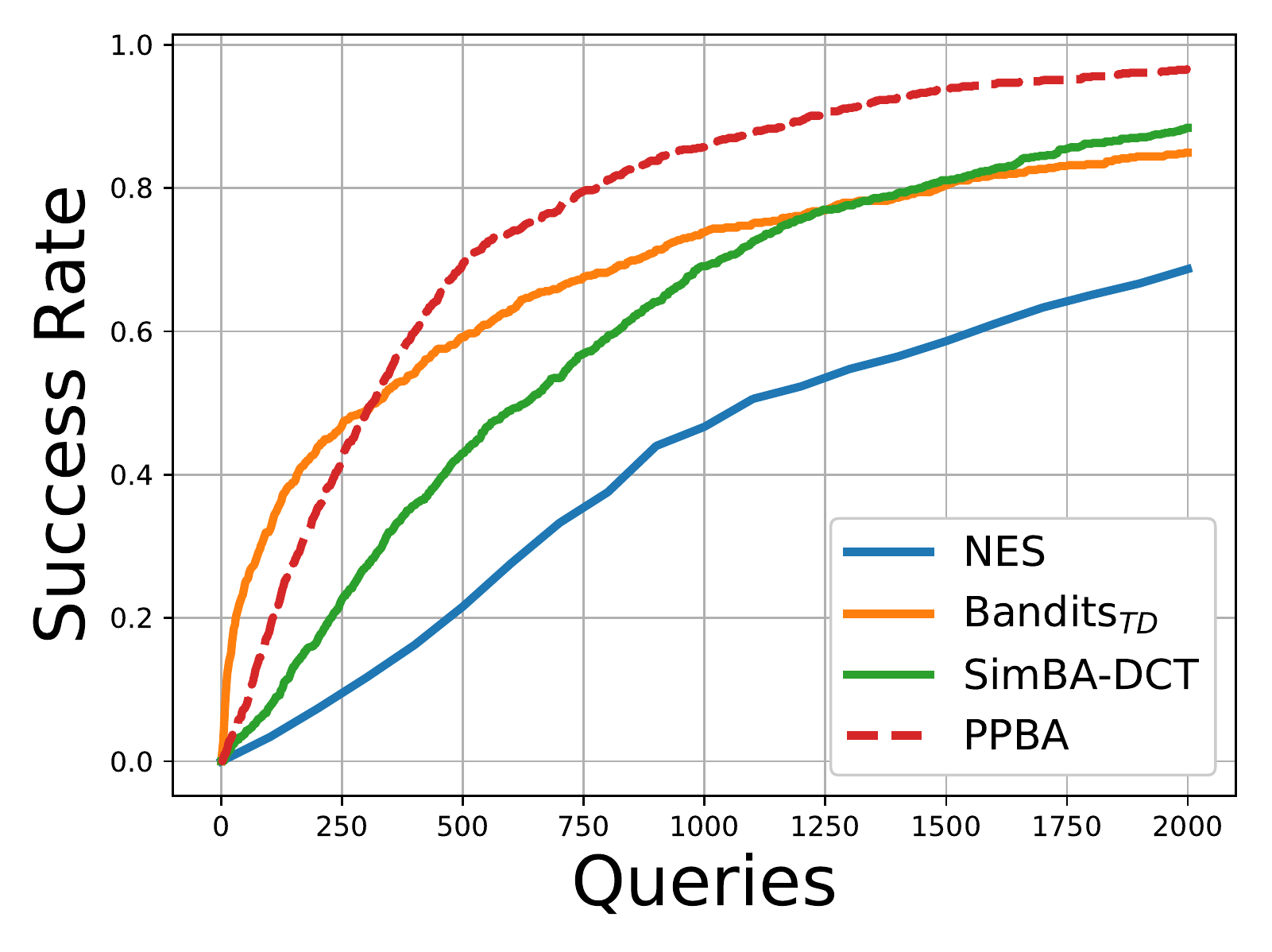} }
        \subfigure[VGG-16]{
        \includegraphics[width=\linewidth]{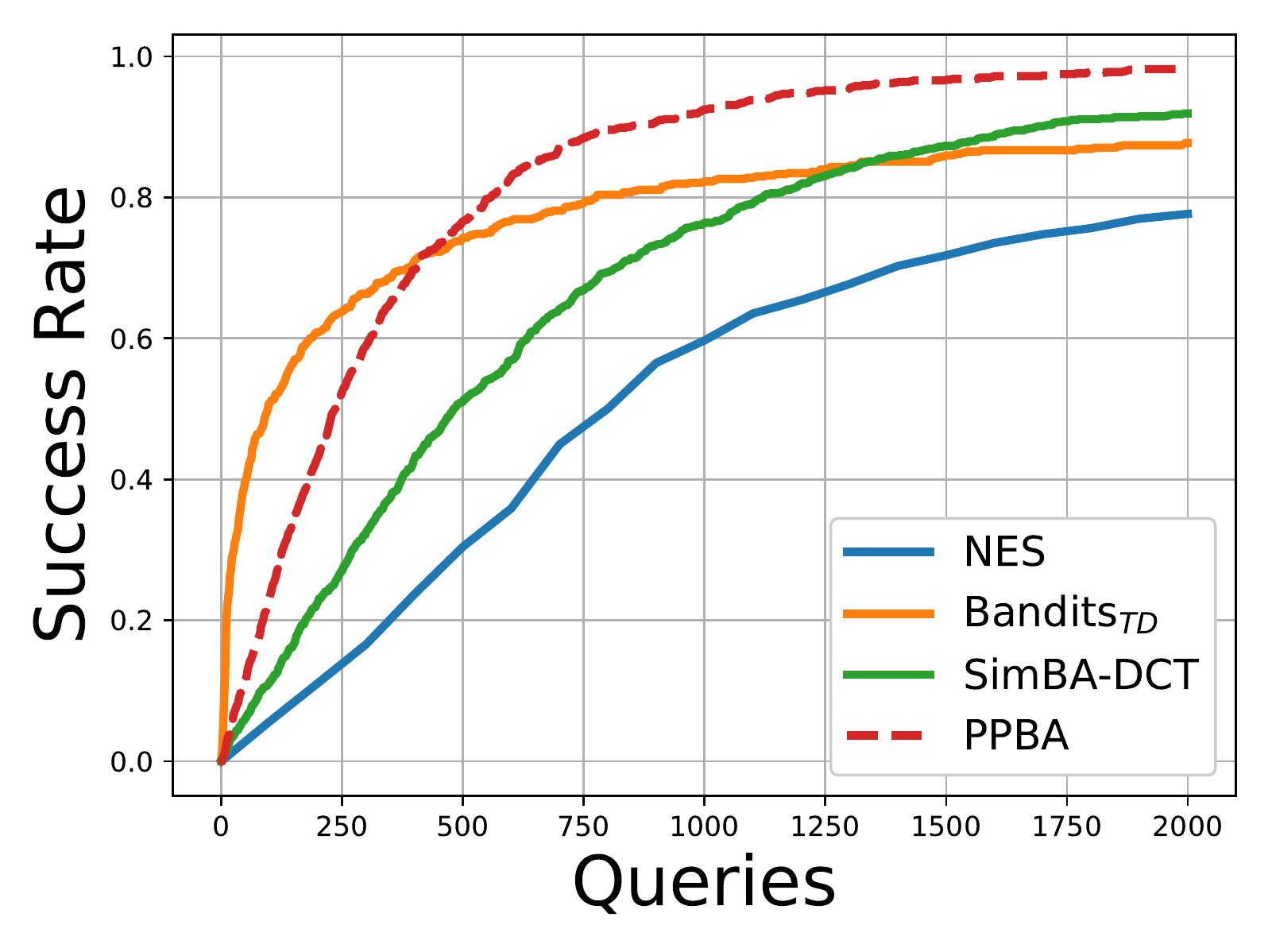} }
        \subfigure[Inception V3]{
        \includegraphics[width=\linewidth]{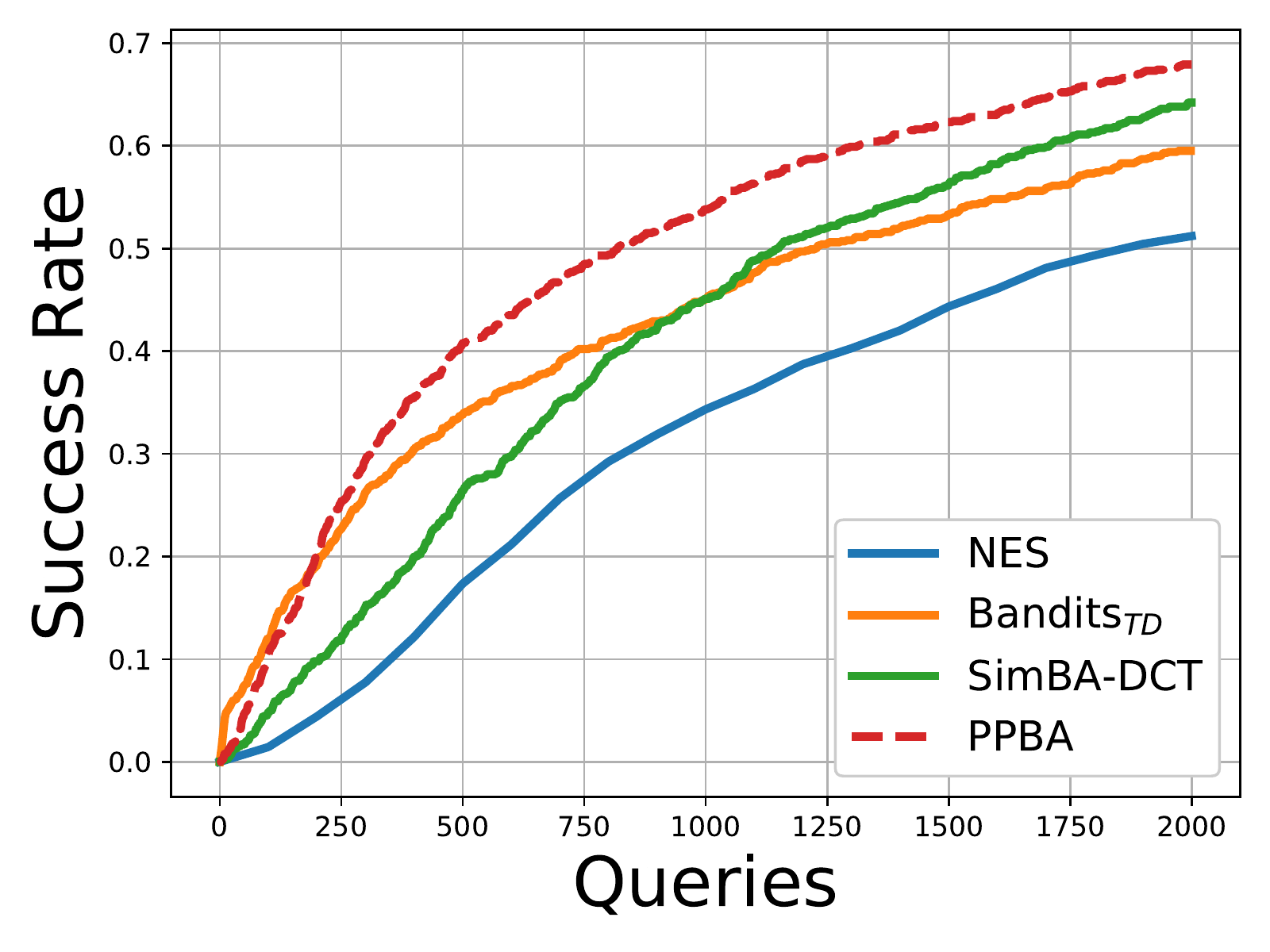} }
    }
    \end{minipage}
    \end{center}
     \vspace{-1em}
    \caption{Curves of the attack success rate versus the number of queries for $l_{\infty}$ attack.}
~\label{fg_inf}
\vspace{-2em}
\end{figure}

\subsection{The Results of PPBA on ImageNet}
We evaluate the performance of our proposed PPBA in Tab.\,\ref{tb_overall} with the maximum $l_2$-norm of perturbation set to $5$ as in~\cite{ilyas2018prior}.
Compared with NES and Bandits$_{TD}$, PRBA and PPBA both achieve a higher success rate with fewer queries.
For example, PRBA improves $12.3\%$ success rate and reduces $24\%$ queries compared with Bandits$_{TD}$ for ResNet50, and PPBA achieves even better results.
Compared with SimBA-DCT, PRBA and PPBA obtain competitive results except for the success rate on ResNet50.
The average queries of PPBA are at most $24\%$ fewer than SimBA-DCT, which makes PPBA standout.
The PPBA method is better than PRBA with a $1.9\%$ higher success rate and $15\%$ fewer queries taking ResNet50 for example, which demonstrates the efficiency of the probability-driven strategy.
To further investigate the relationship between the success rate and the number of queries, we plot the curves of success rate versus queries in Fig.\,\ref{fg_curve}.
From these curves, we conclude that for more samples, PPBA finds feasible solutions within $2,000$ queries more quickly.
The AUC has been calculated for a quantitative comparison in Tab.\,\ref{tb_overall}, which also indicates the superiority of our method.

We evaluate the effectiveness of PPBA under the $l_{\infty}$-norm as well.
Following the setting in~\cite{ilyas2018prior}, the maximum $l_{\infty}$-norm of perturbations is set to $0.05$.
The quantitative results are shown in Tab.\,\ref{tb_inf}, and the curves of the attack success rate versus the number of queries are depicted in Fig.\,\ref{fg_inf}.
Similar to the results of $l_2$ attack, PPBA shows advantages over the baselines with at most $8.2\%$ success rate improved and $26\%$ queries reduced.
We can see that PPBA is more effective and practical enough compared with the state-of-the-art methods.

\begin{figure}[!t]
\vspace{-1em}
    \begin{center}
    \begin{minipage}[t]{0.5\linewidth}
        \centerline{
        \subfigure[ResNet50]{
        \includegraphics[width=\linewidth]{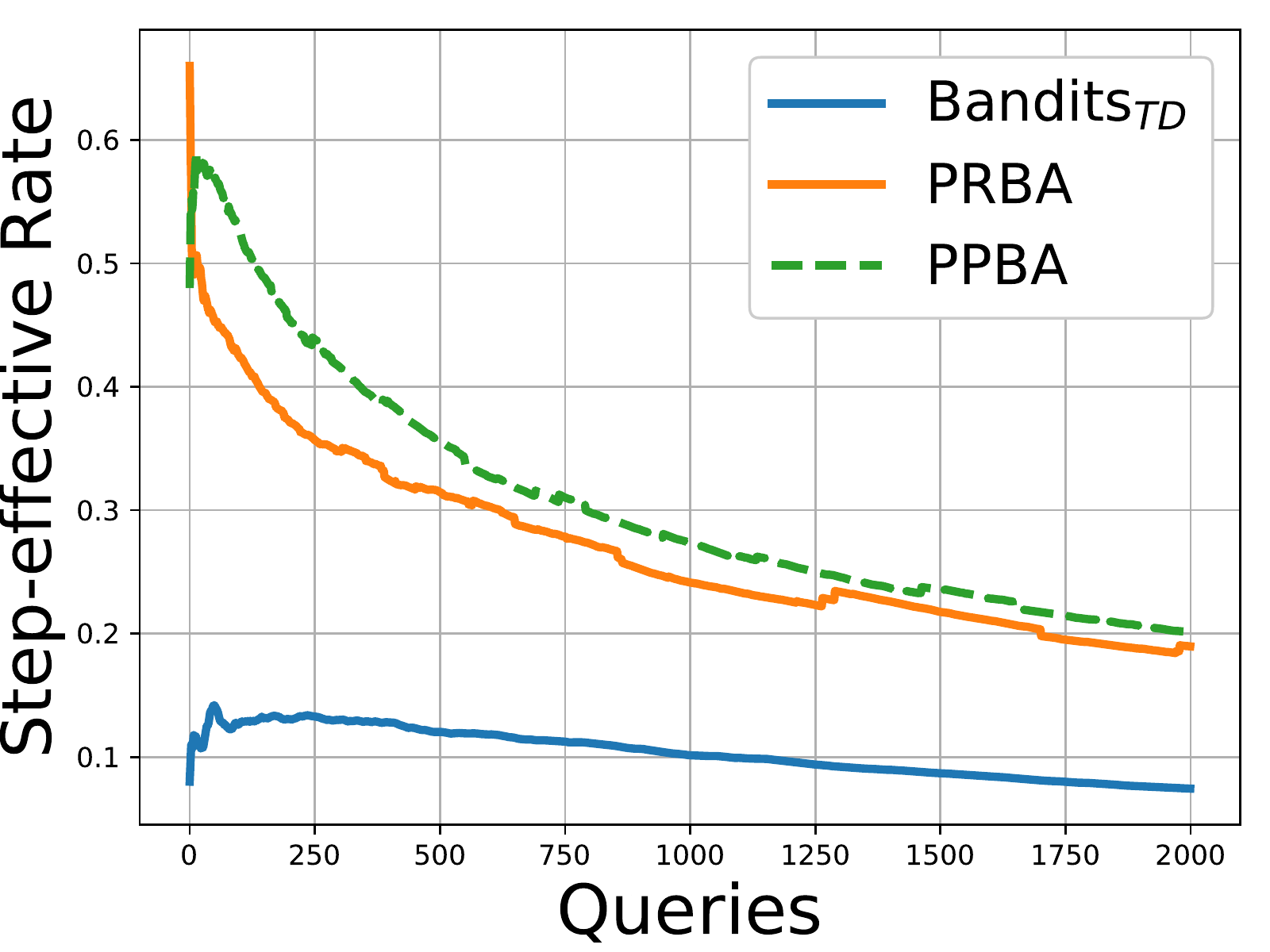} }
        \subfigure[VGG-16]{
        \includegraphics[width=\linewidth]{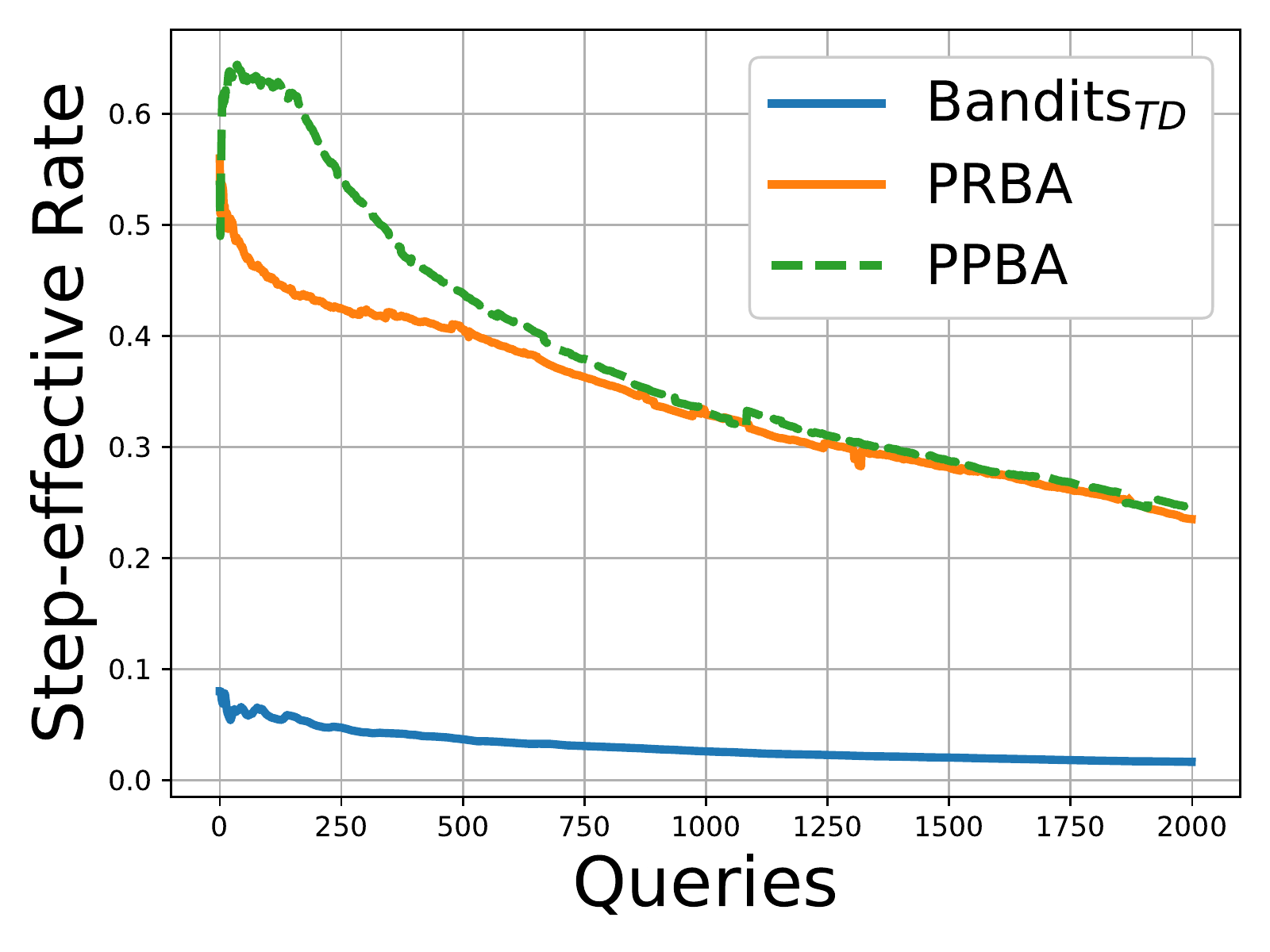} }
    }
    \end{minipage}
    \end{center}
    \caption{Curves of the step-effective rate versus the number of queries for PRBA, PPBA, and Bandits$_{TD}$.}
~\label{fg_eff}
\vspace{-2em}
\end{figure}

To further investigate why PPBA is effective, we measure how often the optimization steps found by the algorithm can bring a descent on the objective function.
Such a step is defined as an effective one, by which the step-effective rate is calculated.
We select 50 images randomly for validation, and plot the curves of the step-effective rate versus the number of queries for PRBA and PPBA, along with Bandits$_{TD}$ in Fig.\,\ref{fg_eff}.
As depicted in the figure, we find that PPBA achieves more than $20\%$ step-effective rate throughout the optimization process, while those of Bandits$_{TD}$ are less than $20\%$.
These results show that the restriction on the optimization step and the probability-driven strategy improve the sample efficiency, and help PPBA find feasible solutions quickly.

\subsection{On Attacking Google Cloud Vision}
\begin{figure}[!t]
    \centering
    \begin{minipage}[t]{0.5\linewidth}
        \centerline{
        \subfigure{
        \includegraphics[width=\linewidth]{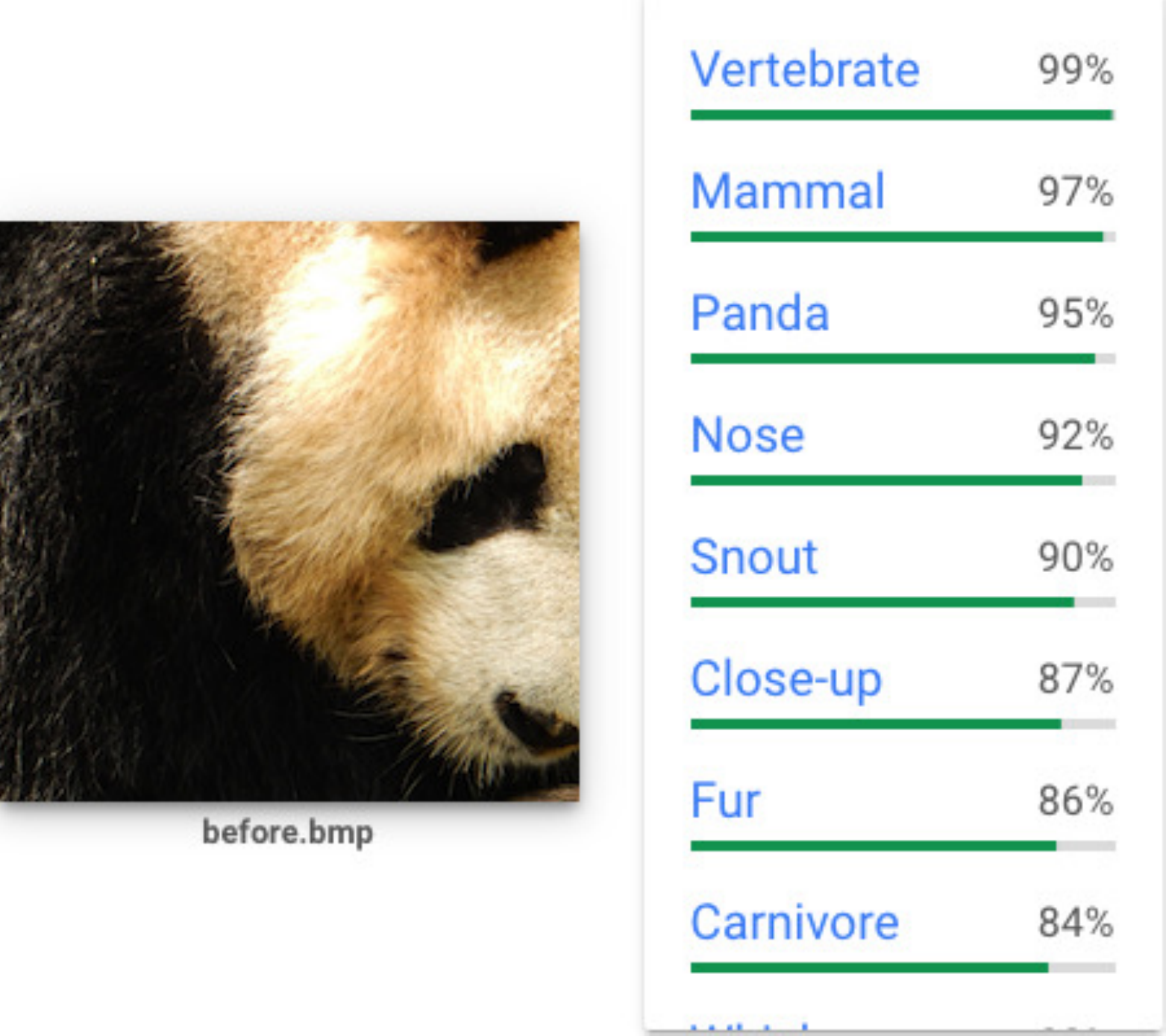} }
        \subfigure{
        \includegraphics[width=\linewidth]{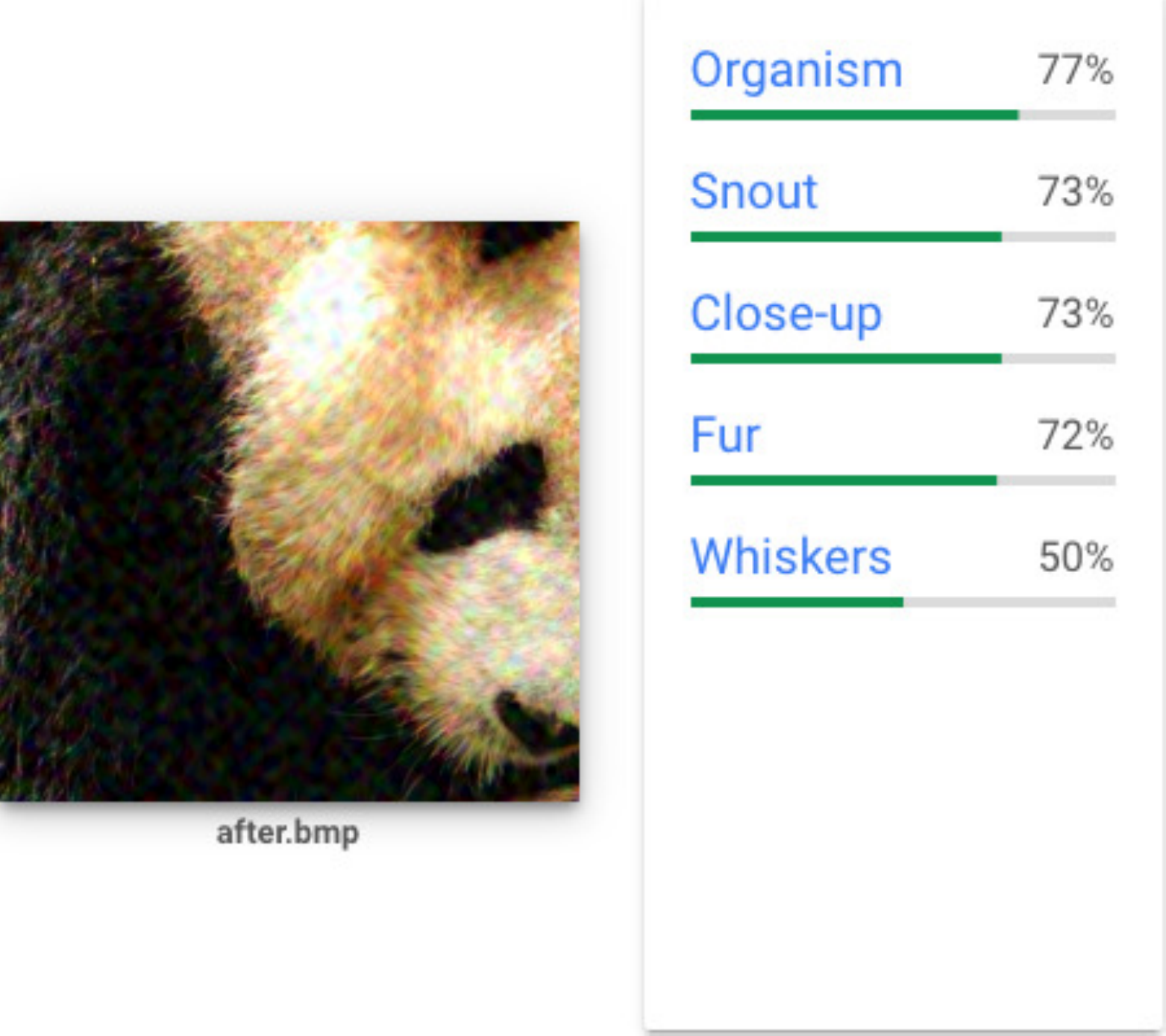} }
    }
    \end{minipage}

    \vspace{-0.7em}
    \hrulefill
    \vspace{-0.2em}

    \begin{minipage}[t]{0.5\linewidth}
        \centerline{
        \subfigure{
        \includegraphics[width=\linewidth]{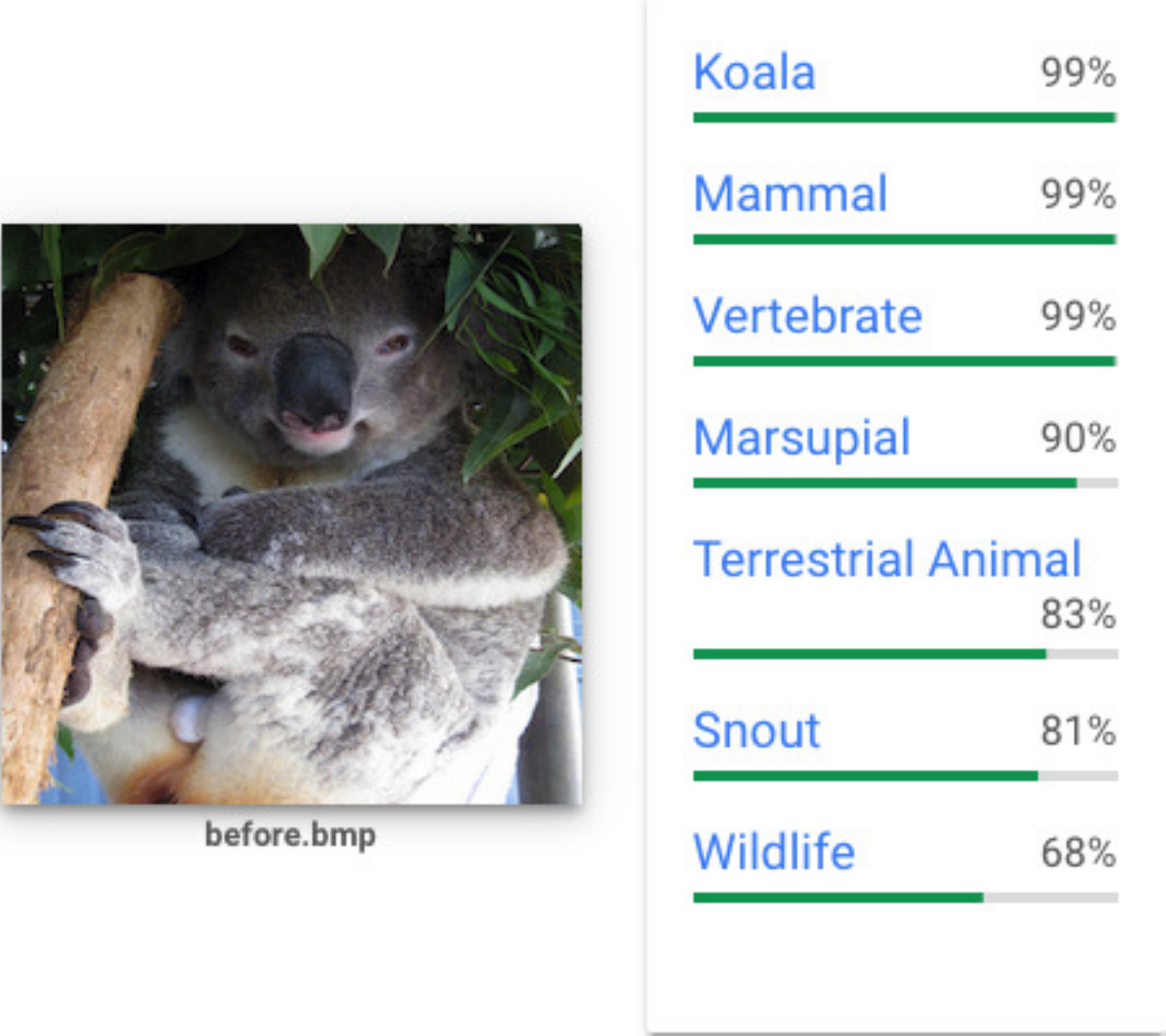} }
        \subfigure{
        \includegraphics[width=\linewidth]{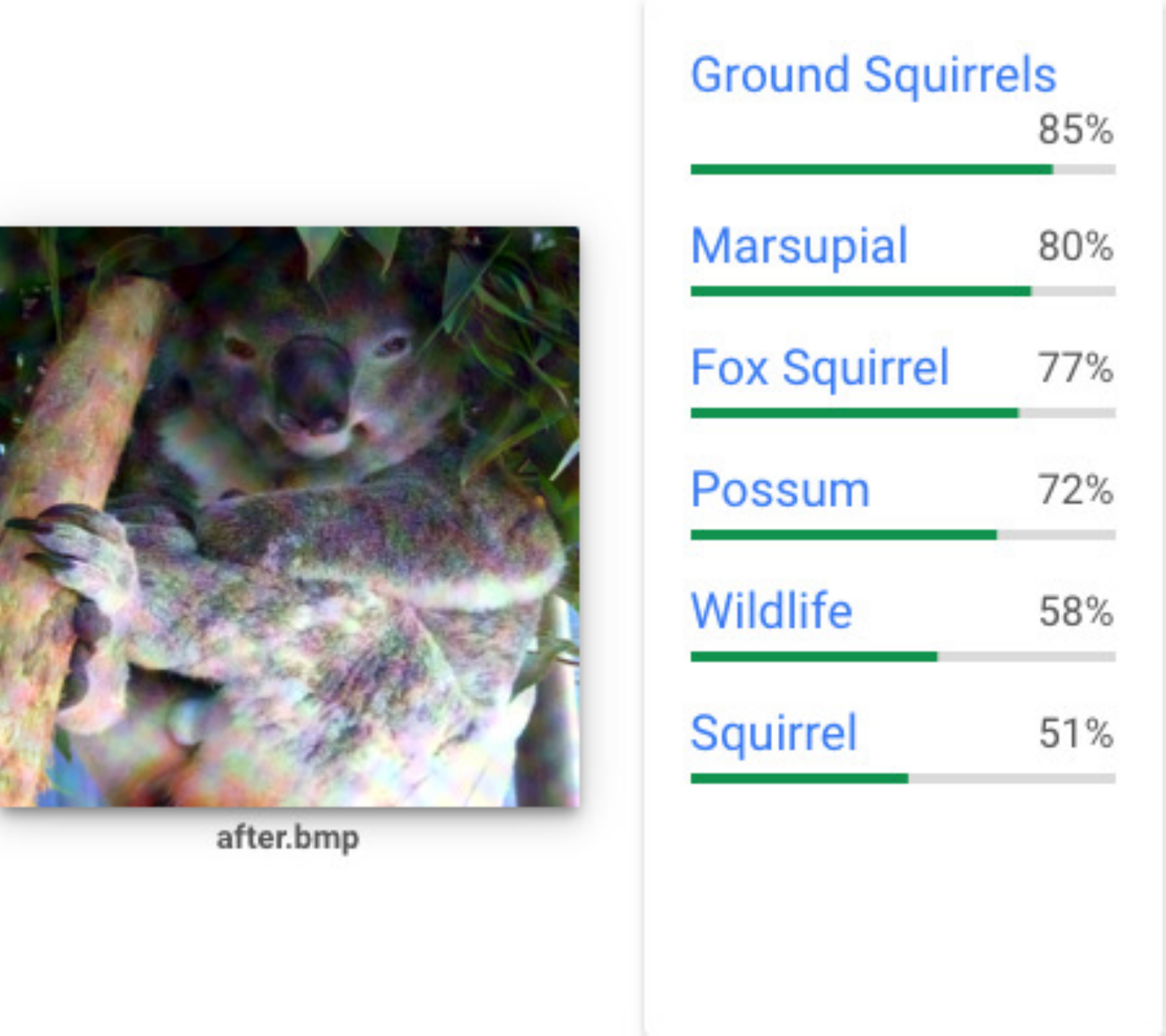} }
    }
    \end{minipage}

\caption{Examples of attacking the Google Cloud Vision API. The images on the left are the original ones and the images on the right are perturbed by PPBA to remove the top-3 labels.
    Taking the first row for example, the Google Cloud Vision API knows the left one is a \emph{panda}, but the right one is not a \emph{panda} anymore.}
~\label{fg_google}
\vspace{-1.5em}
\end{figure}

To demonstrate the effectiveness of our method against real-world online systems, we conduct attacks against the Google Cloud Vision API, which is an online service that offers powerful classification models.
Attacking this system is significantly harder than attacking models pre-trained on ImageNet, since the model is trained on more classes and the exact classes are unclear, while only the top-$k$ labels with their corresponding probabilities for input images can be obtained.
We aim to attack the system by removing the top-$3$ labels presented in the original returned list.
As in~\cite{guo2019simple}, we set the adversarial loss as the maximum of the original top-$3$ labels' returned probabilities,
and minimize the loss with our PPBA.
Fig.\,\ref{fg_google} shows two attack examples. The images on the left are the original ones and the images on the right are perturbed by PPBA.
Taking the first row for example, the top-$3$ labels from the original returned list are related to \emph{panda} with more than $95\%$ probabilities.
After perturbed by PPBA, the concepts related to \emph{panda} disappear from the list, and the Google Cloud Vision API gives the labels related to the local content of the image.

\begin{figure}[!t]
    \begin{center}
        \includegraphics[width=\linewidth]{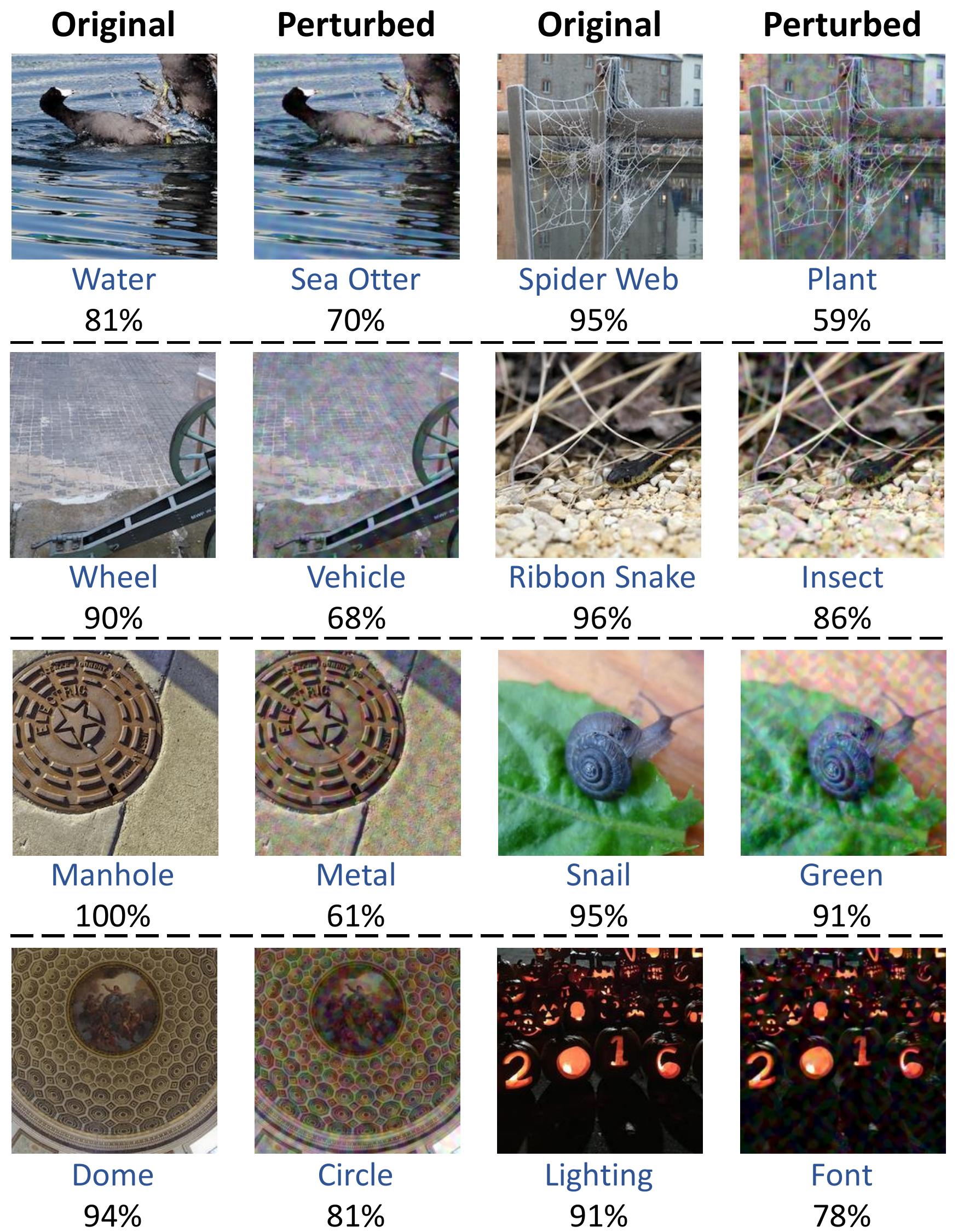}
    \end{center}
\vspace{-1em}
\caption{Visual examples for attacking the Google Cloud Vision API. In each pair, the left is the original image, and the right is the one perturbed by PPBA. The top-$1$ predictions with the probabilities are listed under the images.}
~\label{fg_google_more}
\vspace{-2em}
\end{figure}

Considering the cost the Google Cloud Vision API charges, we evaluate our method on 50 randomly selected images.
We adopt a larger $\rho=0.1$, and set the maximum $l_{\infty}$-norm as $16/255$ ($16/255$ is widely used in recent attack competitions\footnote{\url{https://www.kaggle.com/c/nips-2017-non-targeted-adversarial-attack}, \url{http://hof.geekpwn.org/caad/en/}}).
As a result, PPBA obtains an $84\%$ success rate with $314$ average queries on success samples under this setting, which demonstrates that PPBA is practical for real-world systems.
More visual results are given in Fig.\,\ref{fg_google_more}.

\section{Conclusion}
In this paper, we tackle the problem of the high query budget that black-box attacks suffer.
We propose a novel projection \& probability-driven attack, which mainly focuses on reducing the solution space and improving the optimization.
Towards reducing the solution space, we propose to utilize a low-frequency constrained sensing matrix to reduce the dimensionality of the solution space, inspired by the compressed sensing theory and the low-frequency hypothesis.
Based on the sensing matrix, we further propose a probability-driven optimization that makes the best use of all queries over the optimization process.
We evaluate our proposed method on widely-used neural networks pre-trained on ImageNet, \emph{i.e.}, ResNet50, VGG-16 and Inception V3, in which our method shows significantly higher attack performance with fewer queries compared with the state-of-the-art methods.
Finally, we also attack the real-world system, \emph{i.e.}, Google Cloud Vision API, with a success rate as high as $84\%$, which further demonstrates the practicality of our method.
Last but not least, our work serves as an inspiration in designing more robust models.
We leave it for our future work.

\paragraph{Acknowledgements.}
\footnotesize{
  This work is supported by the Nature Science Foundation of China (No.U1705262, No.61772443, No.61572410, No.61802324 and No.61702136),
  National Key R\&D Program (No.2017YFC0113000, and No.2016YFB1001503),
  and Nature Science Foundation of Fujian Province, China (No. 2017J01125 and No. 2018J01106).
}

{\small
\bibliographystyle{ieee_fullname}
\bibliography{egbib.bib}
}

\end{document}